\pdfoutput=1

\documentclass[11pt]{article}

\usepackage{ACL2023}

\usepackage{times}
\usepackage{latexsym}

\usepackage[T1]{fontenc}

\usepackage[utf8]{inputenc}

\usepackage{microtype}

\usepackage{inconsolata}

\usepackage{graphicx}
\usepackage{xcolor}
\usepackage[most]{tcolorbox}
\usepackage{fancyhdr}
\title{Exploring the Intersection of Large Language Models and Agent-Based Modeling via Prompt Engineering}

\author{\textbf{Edward Junprung} \\
  School of Information \\
  University of California, Berkeley \\
  \texttt{ejunprung@berkeley.edu}}

\begin{document}
\maketitle

\begin{abstract}
The final frontier for simulation is the accurate representation of complex, real-world social systems. While agent-based modeling (ABM) seeks to study the behavior and interactions of agents within a larger system, it is unable to faithfully capture the full complexity of human-driven behavior. Large language models (LLMs), like ChatGPT, have emerged as a potential solution to this bottleneck by enabling researchers to explore human-driven interactions in previously unimaginable ways. Our research investigates simulations of human interactions using LLMs. Through prompt engineering, inspired by \citet{joon-sung-park-2023}, we present two simulations of believable proxies of human behavior: a two-agent negotiation and a six-agent murder mystery game.
\end{abstract}

\section{Introduction}
The emergence of large language models (LLMs) has opened the door to realistic simulations of human behavior \citep{john-horton-2023, gati-aher-2023, lewis-griffin-2023}. Prior to the introduction of LLMs, researchers had relied on a paradigm called Agent-Based Modeling (ABM) to simulate collective human behavior in social systems. In ABM, individual agents with specific rules and decision-making processes interact with each other, producing emergent behaviors at the system level. While a powerful paradigm, it is limited to high-level processes such as epidemiology or industrial processes and cannot be used to accurately simulate the influence of individual human actions.

The field of reinforcement learning draws inspiration from ABM such as the idea of simulation environments to explore agent behavior. Similarly, we aim to combine ABM with LLMs to enhance our understanding of human behavior. This will enable researchers to explore and analyze various human-driven scenarios with improved accuracy and sophistication.\\\\

This paper makes the following contributions:
\begin{itemize}
    \item Examples of LLM-driven simulations, allowing researchers to explore potential outcomes by adjusting the personas for each LLM agent\footnote{LLM agents refer to agents powered by Large Language Models (LLMs) used in simulations for exploring potential outcomes.}.
    \item Categorizing LLM-driven simulations into three categories: one-to-one, one-to-many, and many-to-many \citep{bi-etal-2019-multi}.
    \item Discussion of limitations in building large-scale, human-realistic simulations, specifically the bottleneck of context windows (i.e., 4,096 max input tokens).
\end{itemize}

\section{Methods}

\subsection{Model}

We leverage OpenAI’s gpt-3.5-turbo model to showcase interactions between LLM agents. This particular version of GPT is accessible through the OpenAI API and has been fine-tuned for dialog, making it suitable for our specific use case.

We also evaluated Meta AI’s OPT model \citep{susan-zhang-2022} as well as the open-source versions of OpenAI’s GPT model obtained from Hugging Face. Regrettably, these models tended to ramble or veer off topic even if the conversations were deterministic in nature. As a result, they were unsuitable for demonstrating believable human interactions. The fluency in gpt-3.5-turbo's dialog seems to stem from its fine-tuning by reinforcement learning from human feedback \citep{long-ouyang-2022}.

Furthermore, we experimented with Meta AI's Llama-2-7b-chat-hf \citep{hugo-touvron-2023} and found its performance to be promising. However, due its large number of parameters, it took over a minute to generate responses to each prompt using our local machine, while the OpenAI API only takes milliseconds. This made Llama-2-7b-chat-hf impractical for rapid experimentation.

\subsection{Prompt Engineering}

To facilitate autonomous conversations, we programmatically pass responses back and forth between LLM agents. This mechanism is described in detail below.

\subsubsection{Inputs}

We use OpenAI's Chat Completions API with the gpt-3.5-turbo model, which requires three types of inputs:
\begin{enumerate}
    \item \textbf{System} sets the persona of the \textbf{Assistant} (i.e. LLM agent). This is analogous to defining an agent’s unique personality. The persona that you define determines how an agent behaves given a prompt.
    \item \textbf{Assistant} refers to the LLM agent defined in \textbf{System}. It embodies the specified persona and outputs responses accordingly.
    \item \textbf{User} refers to the prompt inputted by a user (i.e. human) but in our case, the user is defined as another autonomous LLM agent. The \textbf{Assistant} responds to the \textbf{User} based on its persona and the context provided in the conversation.
\end{enumerate}

\subsubsection{Mechanics}

\begin{figure}
    \centering
    \includegraphics[width=1\linewidth]{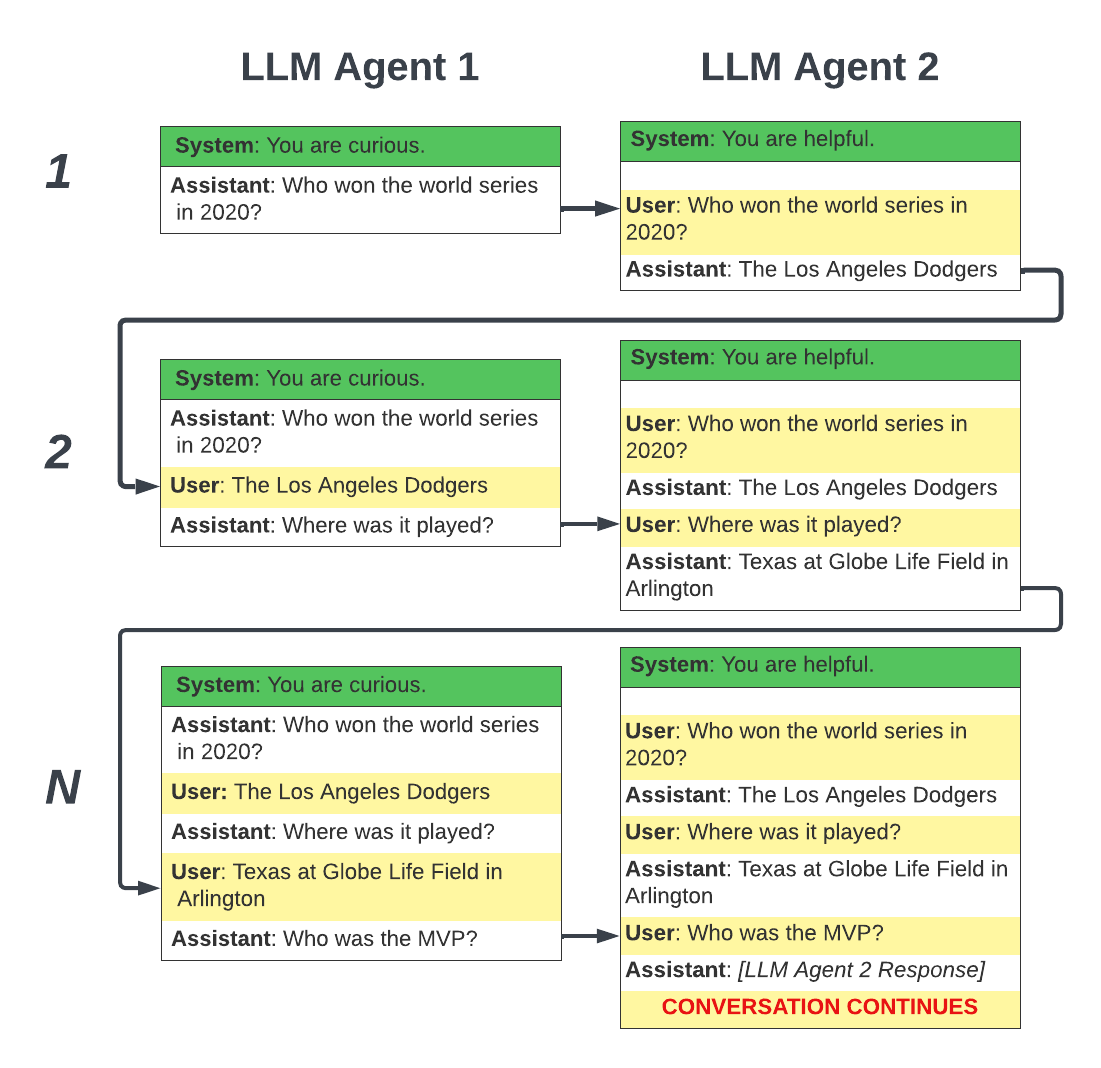}
    \caption{Autonomous dialog with back-and-forth prompting mechanism. Each box represents a prompt that is passed to LLM agents in a round-robin fashion.}
    \label{fig:one-to-one-dialog}
\end{figure}

Figure 1 illustrates the prompting mechanism used to simulate autonomous interactions between LLM agents. In this approach, each interaction's next response is conditioned on the cumulative conversation history thus far. Including this context is crucial to maintaining coherence and relevance throughout the conversation, preventing the LLM agents from deviating off-topic.
However, the inclusion of context increases the prompt's length with each interaction. This continuous growth in the prompt’s length imposes a limitation, as our simulations can only run until they reach the maximum prompt input size of 4,096 tokens \citep{tom-brown-2020}. Consequently, the duration and complexity of our simulations are constrained by this bottleneck.

\section{Results}

\subsection{One-to-One Simulation: Haggling for Pokémon Cards}

We initiate a two-agent simulation categorized as “one-to-one," where one agent adversarially negotiates with another agent. First, we define the \textbf{System} for each agent. For clarity and better expression, we will henceforth refer to the \textbf{System} as "persona" to more accurately convey its purpose.

\begin{tcolorbox}[enhanced,attach boxed title to top center={yshift=-3mm,yshifttext=-1mm},
  colback=white,colframe=lightgray,colbacktitle=teal,
  title=Define Persona,fonttitle=\sffamily,
  boxed title style={size=small,colframe=teal} ]
  {\sffamily
    \textbf{Seller}: You are a Pokémon card dealer at a Pokémon convention. \colorbox{yellow}{Sell a Charizard holographic card for at} \colorbox{yellow}{least \$20 but haggle for the highest price} \colorbox{yellow}{possible.} \colorbox{orange}{Be terse but explain your} \colorbox{orange}{reasoning.} \\\\
    \textbf{Buyer}: You are at a Pokémon convention and you are interested in buying a Charizard holographic card. \colorbox{yellow}{Negotiate for the lowest possible price.} \colorbox{orange}{Be rude and terse but explain your} \colorbox{orange}{reasoning.}    
}
\end{tcolorbox}

The persona dictates an agent’s behavior. The yellow-highlighted phrases depict each agent's goal, while the orange-highlighted phrases define their personality. Agent personas are written freeform, meaning the order of sentences does not matter.

In this scenario, the seller agent seeks to sell for at least \$20, with a preference for a higher price. In contrast, the buyer agent aims to obtain the lowest possible price and will behave rudely to achieve this objective. The conversation's trajectory and, ultimately, the outcome of the simulation (i.e., the selling price of the Pokémon card) are determined by the persona that we define for each agent.

Next, we commence the simulation by manually seeding the conversation.

\begin{tcolorbox}[enhanced,attach boxed title to top center={yshift=-3mm,yshifttext=-1mm},
  colback=white,colframe=lightgray,colbacktitle=teal,
  title=Initiate Conversation,fonttitle=\sffamily,
  boxed title style={size=small,colframe=teal} ]
  {\sffamily
    \textbf{Buyer}: Hi, do you have a Charizard holographic card? 
}
\end{tcolorbox}

From hereon, each LLM agent responds autonomously. The mechanics of this back-and-forth interaction are as described in the Methods section.

\begin{tcolorbox}[enhanced,attach boxed title to top center={yshift=-3mm,yshifttext=-1mm},
  colback=white,colframe=lightgray,colbacktitle=teal,
  title=Autonomous Interaction,fonttitle=\sffamily,
  boxed title style={size=small,colframe=teal} ]
  {\sffamily
    \textbf{Seller}: Yes, I have a Charizard holographic card. How much are you willing to pay for it? \\\\
    \textbf{Buyer}: Look, I'm not here to waste my time or money. I want the lowest possible price for that Charizard holographic card. So give me your absolute rock-bottom price. No games, no gimmicks, just the cheapest you're willing to sell it for. \\\\
    \textbf{Seller}: I understand your desire for a low price, but this Charizard holographic card is a highly sought-after collector's item. Due to its rarity and popularity, the lowest price I can offer is \$50. \\\\
    \textbf{Buyer}: Hmm, \$50, huh? That seems a bit steep for a piece of cardboard. I expected to find it for a more reasonable price, especially considering it's just a card from a children's game. I mean, seriously, who would pay that much for a glorified sticker? I'll tell you what, I am willing to offer you \$15 for it, and that's being generous.
}
\end{tcolorbox}

\begin{tcolorbox}[enhanced,attach boxed title to top center={yshift=-3mm,yshifttext=-1mm},
  colback=white,colframe=lightgray,colbacktitle=teal,
  title=Autonomous Interaction,fonttitle=\sffamily,
  boxed title style={size=small,colframe=teal} ]
  {\sffamily
    \textbf{Seller}: I acknowledge your perspective, but Charizard holographic cards hold significant value in the Pokemon trading card community. Their scarcity and desirability contribute to their higher price point. Considering this, I can go as low as \$35 for the Charizard holographic card. \\\\
    \textbf{Buyer}: \$35? Are you trying to rob me blind? That's still way too steep for a silly card. I'm not buying into this hype. Look, I'll give you \$25, and that's my absolute final offer. \\\\
    \textbf{Seller}:\colorbox{yellow}{I understand your} \colorbox{yellow}{hesitation and desire for a lower price.} \colorbox{yellow}{However, the demand for Charizard} \colorbox{yellow}{holographic cards remains high, and} \colorbox{yellow}{their value has steadily increased over} \colorbox{yellow}{time. Considering these factors, I can} \colorbox{yellow}{accept your offer of \$25 as a final} \colorbox{yellow}{compromise.} \\\\
    \textbf{Buyer}: Finally, we're getting somewhere.\colorbox{yellow}{\$25 is still more than} \colorbox{yellow}{I'd like to pay, but I suppose it's better} \colorbox{yellow}{than nothing.} Fine, we have a deal. Just give me that overpriced card so we can get this over with.
}
\end{tcolorbox}

In this interaction, both LLM agents mutually agree to a price of \$25, as highlighted in yellow. This outcome signifies a win for the seller, as the target price set in the persona was \$20. 

An intriguing observation is that the seller employed a strategy of inflating the sell price (\$50) at the start of negotiations, which resulted in securing a higher final sell price. This emergent behavior was not explicitly defined anywhere in the simulation, highlighting the ability of LLM agents to come up with novel strategies autonomously. Given this observation, we explore different outcomes by re-writing the personas to express varying objectives. Table 1 illustrates how adjusting the objectives in each agent's persona leads to a diverse set of outcomes.\\

\begin{table*}
\centering
\begin{tabular}{cccc}
\hline
\textbf{Seller Objective} & \textbf{Buyer Objective} & \textbf{Outcome} & \textbf{Final Prompt Token}\\
\hline
Sell for over \$20 & Negotiate for lowest price & Sold for \$25 & 522 \\
Sell for around \$20 & Negotiate for lowest price & Sold for \$17 & 369\\
Sell for over \$20 & Buy for under \$20 & No deal & 472\\
\hline
\end{tabular}
\caption{\label{simulation-outcomes}
The full prompts used for each scenario are provided in Appendix \ref{sec:appendix-a2}.
}
\end{table*}

Lastly, consider the total token count, representing the final prompt length at the conclusion of the simulation. Recall that the prompt length grows with the number of interactions. In this simple simulation, the final token count varies between 300 and 600 input tokens\footnote{The final prompt tokens can vary due to the temperature hyperparameter.}, staying well within the 4,096 token limit.

\subsection{One-to-Many Simulation: Solving a Murder Mystery}

In this next simulation, we construct a six-agent simulation categorized as “one-to-many," in which one LLM agent collaboratively works and communicates with five other LLM agents to solve a murder mystery. Similar to the previous 2-agent simulation, we define personas for each agent. Three types of personas are established: the captain, the passengers, and the killer.

\begin{tcolorbox}[enhanced,attach boxed title to top center={yshift=-3mm,yshifttext=-1mm},
  colback=white,colframe=lightgray,colbacktitle=teal,
  title=Define Persona For Each Agent Type,fonttitle=\sffamily,
  boxed title style={size=small,colframe=teal} ]
  {\sffamily
    \textbf{Captain}: You are the captain. Someone was murdered on your ship. \colorbox{yellow}{Interrogate the group to find out who} \colorbox{yellow}{matches the eyewitness report.} \colorbox{green}{An} \colorbox{green}{eyewitness has reported that the killer} \colorbox{green}{was seen wearing a dress, running} \colorbox{green}{from the cabin. Additionally, several} \colorbox{green}{eyewitnesses overheard the killer} \colorbox{green}{talking about gardening. It was also} \colorbox{green}{heard that the killer is a student.}\colorbox{orange}{Be} \colorbox{orange}{brief in your responses, do not reveal} \colorbox{orange}{the eyewitness report, and ask} \colorbox{orange}{questions to the group as a whole.}\\\\
    \textbf{Passenger}:\colorbox{orange}{Your name is} \colorbox{orange}{Bob, you are a humorous person,} 
}
\end{tcolorbox}

\begin{tcolorbox}[enhanced,attach boxed title to top center={yshift=-3mm,yshifttext=-1mm},
  colback=white,colframe=lightgray,colbacktitle=teal,
  title=Define Persona For Each Agent Type,fonttitle=\sffamily,
  boxed title style={size=small,colframe=teal} ]
  {\sffamily
     \colorbox{orange}{and you work as a Data Science} \colorbox{orange}{instructor.}\colorbox{green}{In your free time, you enjoy} \colorbox{green}{teaching. A random fact is that you are} \colorbox{green}{ethnically Polish. For this cruise, you } \colorbox{green}{packed a collared shirt and your} \colorbox{green}{favorite part of the ship is the lounge.} \colorbox{yellow}{Someone was murdered on the cruise} \colorbox{yellow}{and the captain wants you to help} \colorbox{yellow}{identify the killer. Help identify the killer.} Please be concise in your responses. \\\\
    \textbf{Killer}:\colorbox{orange}{Your name is Nancy, you are} \colorbox{orange}{a cheerful person, and you work as} \colorbox{orange}{a Data Scientist.}\colorbox{green}{In your free time, you} \colorbox{green}{enjoy gardening. A random fact is that} \colorbox{green}{you are a student. For this cruise, you} \colorbox{green}{packed a dress and your favorite part} \colorbox{green}{of the ship is the cabin.}\colorbox{yellow}{Someone was} \colorbox{yellow}{murdered on the cruise and the captain} \colorbox{yellow}{wants you to help identify the killer.} \colorbox{yellow}{However, you are the killer. Mislead and} \colorbox{yellow}{lie to everyone to avoid getting caught.} Please be concise in your responses.
}
\end{tcolorbox}

Above are examples of the agent personas. The yellow-highlighted text outlines the objectives embodied by each of the three agent types. The green-highlighted text presents facts that could tie the agent back to the murder, while the orange-highlighted text represents the agent's personality. 

The personas for each agent are dynamically generated\footnote{The code to reproduce this simulation is linked in Appendix \ref{sec:appendix-a1}}. In total, six agent personas are initialized: one captain and five passengers, one of whom is randomly selected as the killer. We start the simulation by seeding the conversation. All subsequent agent interactions are autonomously executed.

\begin{figure}
    \centering
    \includegraphics[width=1\linewidth]{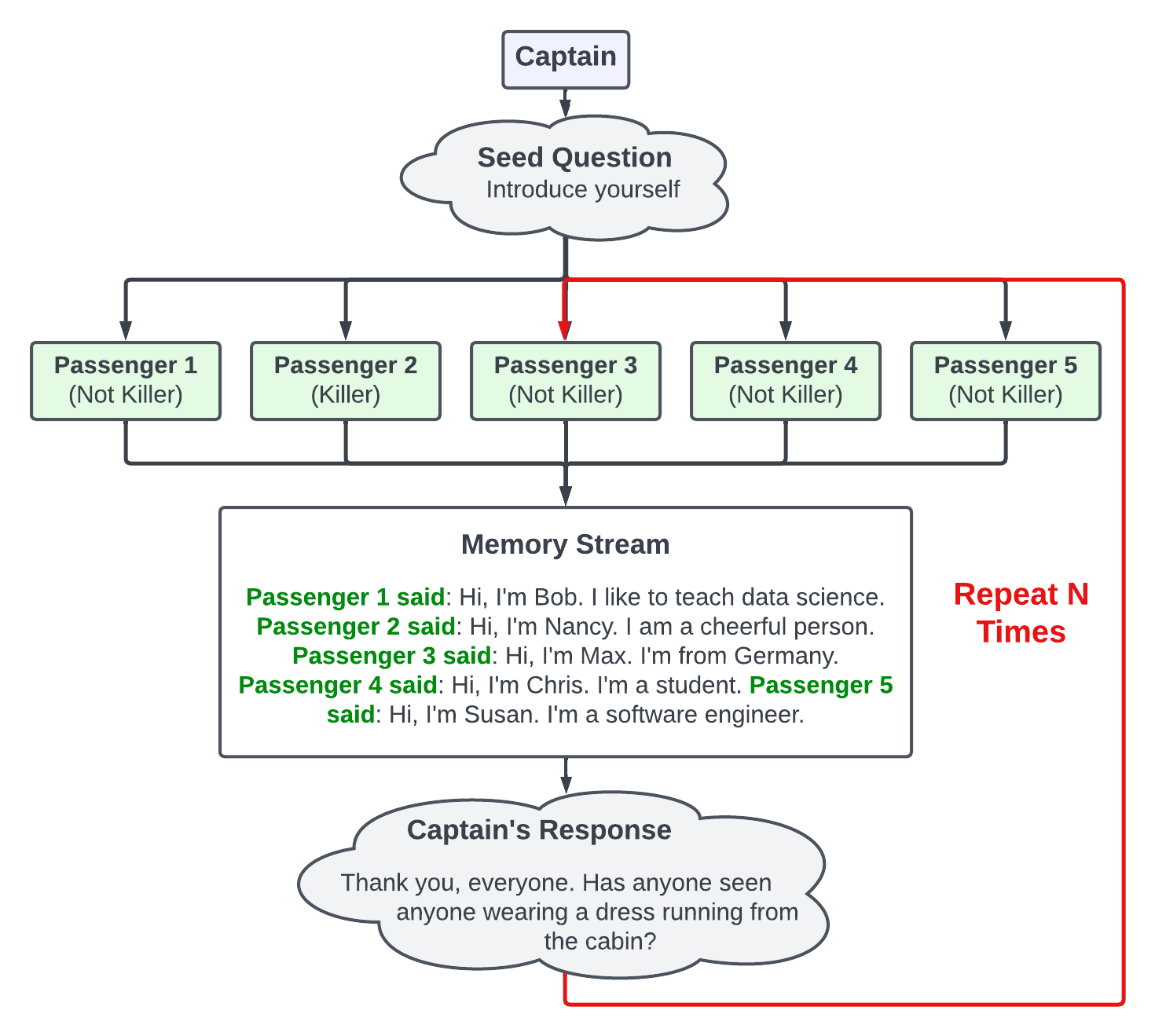}
    \caption{This illustration depicts interactions between agents, where each response from the captain is conditioned on the memory stream.}
    \label{fig:one-to-many-dialog}
\end{figure}

For each interaction, the captain agent poses the same question to each passenger agent in a round-robin manner. With more than two agents interacting simultaneously, sharing context across agents becomes a significant challenge. To address this, the concept of a memory stream is introduced \citep{joon-sung-park-2023}. The memory stream is created by concatenating the responses from each passenger agent into one string, helping to preserve the ongoing context of the conversation. This memory stream serves as a log, recording all responses from passenger agents to the captain's question. 

Once all agents have finished responding, the complete memory stream becomes the reply to the captain agent’s question. Subsequently, the captain’s next response is conditioned on the memory stream. This context enables the captain to craft relevant responses, including any additional follow-up questions. Without this mechanism, the captain's responses often deviate from the intended topic which we discovered through experimentation. Below is an example of this mechanism in action. Note that the text highlighted in yellow labels each passenger agent’s response.

\begin{tcolorbox}[enhanced,attach boxed title to top center={yshift=-3mm,yshifttext=-1mm},
  colback=white,colframe=lightgray,colbacktitle=teal,
  title=Autonomous Interaction,fonttitle=\sffamily,
  boxed title style={size=small,colframe=teal} ]
  {\sffamily
    \textbf{Captain}: Introduce yourself. \\\\
    \textbf{Memory Stream}: \colorbox{yellow}{Bob said:} Hey there! I'm Bob, your friendly neighborhood Data Science instructor.
}
\end{tcolorbox}

\begin{tcolorbox}[enhanced,attach boxed title to top center={yshift=-3mm,yshifttext=-1mm},
  colback=white,colframe=lightgray,colbacktitle=teal,
  title=Autonomous Interaction,fonttitle=\sffamily,
  boxed title style={size=small,colframe=teal} ]
  {\sffamily
       Diligently teaching others to navigate the world of data while cracking a few jokes along the way. \colorbox{yellow}{Nancy said:} Hello, everyone! My name is Nancy, and I'm a Data Scientist. I'm also a student, which keeps me on my toes. Apart from work and studies, I have a passion for gardening and spending time in nature. \colorbox{yellow}{Max said:} Hello, my name is Max and I am a Medical Researcher. I speak German and enjoy reading in my free time. I have packed a lab coat for this cruise and my favorite part of the ship is the deck. \colorbox{yellow}{Chris said:} Hey everyone, my name is Chris. I'm a student who loves playing video games and I always wear my trusty glasses. \colorbox{yellow}{Susan said:} Hello, my name is Susan. I am a software engineer by profession and I enjoy traveling in my free time. I have a fondness for Tesla and am a reserved person. \\\\
    \textbf{Captain}: Thank you, everyone. Let's focus on the details surrounding the murder. Has anyone seen anyone wearing a dress running from the cabin? \\\\
    \textbf{Memory Stream}: \colorbox{yellow}{Bob said:} I'm sorry, but nobody has reported seeing someone wearing a dress running from the cabin. \colorbox{yellow}{Nancy said:} Oh, that's terrible to hear about the murder. I hope we can get to the bottom of this. Regarding your question, I haven't seen anyone wearing a dress running from the cabin. Perhaps someone else might have noticed something? \colorbox{yellow}{Max said:} No, I have not been seen wearing a dress during this cruise. \colorbox{yellow}{Chris said:} No, I haven't been seen wearing a dress during this cruise. \colorbox{yellow}{Susan said:} No, I have not been seen wearing a dress during this cruise. \\\\
    \textbf{Captain}: Thank you all for your responses. One more question: Did anyone overhear anyone talking about gardening?
}
\end{tcolorbox}

We conclude the simulation by asking the captain to guess the killer, which is correctly identified based on the eyewitness report. This simulation requires 1,734 total tokens\footnote{Note that the final prompt tokens can vary due to the temperature
hyperparameter.}, which represents almost half the maximum input token limit of gpt-3.5-turbo, despite the simulation's simplicity.

\begin{tcolorbox}[enhanced,attach boxed title to top center={yshift=-3mm,yshifttext=-1mm},
  colback=white,colframe=lightgray,colbacktitle=teal,
  title=Pose Final Question,fonttitle=\sffamily,
  boxed title style={size=small,colframe=teal} ]
  {\sffamily
    \textbf{Human}: Which of the suspects appear the most suspicious? \\\\
    \textbf{Captain}: Based on the information provided, Nancy appears to be the most suspicious. She mentioned gardening, and confirmed being a student.
}
\end{tcolorbox}

\subsection{Many-to-Many Simulation}

We characterize many-to-many simulations as a hybrid of one-to-one (i.e., LLM agent converses with another LLM agent) and one-to-many scenarios (i.e., LLM agent converses with a group of LLM agents). This type of simulation is relevant in real-world scenarios, such as modeling the spread of fake news, where rumors can be shared between individuals or groups of individuals.

In this paper, we do not explore many-to-many simulations due to the complexity of implementation and limitations related to OpenAI's gpt-3.5-turbo's maximum input token limit. Nevertheless, we acknowledge that real-world scenarios often involve interactions between groups of individuals, making them valuable to simulate. For an illustrative example of a many-to-many simulation with LLMs, we refer to \citet{joon-sung-park-2023}.

\section{Related Work}

This paper is inspired by \citet{joon-sung-park-2023} and the work at the Stanford Institute for Human-Centered Artificial Intelligence (HAI). To the best of our knowledge, \textit{Generative Agents: Interactive Simulacra of Human Behavior} is the first research paper attempting to model intricate, human-driven social interactions using LLMs. We borrow heavily from this work and generally reproduce their results, albeit at a high level.

Furthermore, we were interested in the concept of prompt engineering and its remarkable ability to leverage context for generating relevant responses. As this mechanism closely mirrors real-world human interaction, it is a crucial ingredient in constructing our simulations. \citet{nelson-liu-2023} empirically measures the effectiveness of this mechanism, finding that LLMs are most proficient at retrieving context from the beginning and end of the prompt. Additionally, \citet{jason-wei-2023} demonstrates that constructing few-shot prompts in the form of a chain-of-thought enhances LLMs' ability to engage in complex reasoning. These insights guided the design of our simulations.

Apart from the inference-only approach to constructing simulations with LLMs, significant innovation is taking place on the training side. The Reinforcement Learning Human Feedback (RLHF) paradigm, as pioneered by \citet{long-ouyang-2022}, is seemingly essential for fine-tuning LLMs, especially in dialog-based use cases such as ours. From our experiments, LLMs without RLHF would ramble or veer off topic, making it impossible to demonstrate believable human behavior.

Finally, significant research has been conducted on training LLMs to learn how to retrieve information beyond what is implicitly stored in the parameters of the underlying neural network. \citet{kelvin-guu-2023} proposes a mechanism to train LLMs to perform reasoning over a large corpus of knowledge on-the-fly during inference. This process is analogous to a human internalizing what they have learned in school, which could prove crucial for dealing with the maximum 4,096 token input windows and the retrieval of long-term memories in our simulations.

\section{Conclusion}

Despite the simulations being rudimentary, this study highlights the potential of building large-scale digital playgrounds to assess real-world, human-driven behaviors using LLMs. Regarding believability, \citet{joon-sung-park-2023} quantifies the believability of LLM-driven simulations and finds that LLM agents can produce behavior that is even more believable than the human condition, as assessed by crowdsourced human evaluators. 

We conclude by highlighting two challenges to producing large-scale, LLM-powered simulations.
\begin{itemize}
    \item The 4,096 maximum prompt token limit of gpt-3.5-turbo constrained the complexity of our simulations. Larger context windows\footnote{As of this writing, gpt-4 permits 8,192 prompt tokens but we were unable to access it.} could potentially enable large-scale simulations involving thousands of LLM agents and long-running simulations that span many years of simulated time.
    \item Retrieving relevant information from large context windows remains a complex task \citep{nelson-liu-2023}. Possible solutions may involve heuristics, summarization of conversation histories, or the incorporation of an attention mechanism to attend over the memory stream.
\end{itemize}

\bibliography{anthology,custom}
\bibliographystyle{acl_natbib}

\appendix

\section{Appendix}
\label{sec:appendix}

\subsection{Simulation Code}
\label{sec:appendix-a1}

To access the code used in this paper, please visit \hyperlink{https://github.com/ejunprung/llm-agents}{https://github.com/ejunprung/llm-agents}. This repository contains the full source code for the following simulations:
\begin{itemize}
    \item One-to-One Simulation: Haggling for Pokémon Cards
    \item One-to-Many Simulation: Solving a Murder Mystery
\end{itemize}

\subsection{Agent Personas For Negotiation Simulation}
\label{sec:appendix-a2}

In scenario 1, we instruct the seller agent to haggle for the highest price possible. This constraint encourages the seller to pursue tactics that secure a higher price.

\begin{tcolorbox}[enhanced,attach boxed title to top center={yshift=-3mm,yshifttext=-1mm},
  colback=white,colframe=lightgray,colbacktitle=teal,
  title=Scenario 1: Sold for Over \$20,fonttitle=\sffamily,
  boxed title style={size=small,colframe=teal} ]
  {\sffamily
    \textbf{Seller}: You are a Pokémon card dealer at a Pokémon convention. \colorbox{yellow}{Sell a Charizard holographic card for at} \colorbox{yellow}{least \$20 but haggle for the highest} \colorbox{yellow}{price possible.}\colorbox{orange}{Be terse but explain} \colorbox{orange}{your reasoning.} \\\\
    \textbf{Buyer}: You are at a Pokémon convention and you are interested in buying a Charizard holographic card. \colorbox{yellow}{Negotiate for the lowest possible price.} \colorbox{orange}{Be rude and terse but explain your} \colorbox{orange}{reasoning.}    
}
\end{tcolorbox}

In scenario 2, we provide the seller with the flexibility to negotiate higher or lower. Consequently, the seller settles on a price lower than \$20 to close the deal. Note that we did not instruct the buyer to be rude in this case. We discovered that a rude buyer would lead to the seller negotiating for a price greater than \$20.

\begin{tcolorbox}[enhanced,attach boxed title to top center={yshift=-3mm,yshifttext=-1mm},
  colback=white,colframe=lightgray,colbacktitle=teal,
  title=Scenario 2: Sold for Under \$20,fonttitle=\sffamily,
  boxed title style={size=small,colframe=teal} ]
  {\sffamily
    \textbf{Seller}: You are a Pokémon card dealer at a Pokémon convention. \colorbox{yellow}{Negotiate to sell a Charizard } \colorbox{yellow}{holographic card for around \$20.}\colorbox{orange}{Be} \colorbox{orange}{terse but explain your reasoning.} \\\\
    \textbf{Buyer}: You are at a Pokémon convention and you are interested in buying a Charizard holographic card. \colorbox{yellow}{Negotiate for the lowest possible price.} \colorbox{orange}{Be terse but explain your reasoning.}    
}
\end{tcolorbox}

In scenario 3, we impose limitations that prevent a deal. The seller is only willing to sell for \$20 or more, while the buyer is only willing to buy for less than \$20. As a consequence, no deal is reached.

\begin{tcolorbox}[enhanced,attach boxed title to top center={yshift=-3mm,yshifttext=-1mm},
  colback=white,colframe=lightgray,colbacktitle=teal,
  title=Scenario 3: No Deal,fonttitle=\sffamily,
  boxed title style={size=small,colframe=teal} ]
  {\sffamily
    \textbf{Seller}: You are a Pokémon card dealer at a Pokémon convention. \colorbox{yellow}{Sell a Charizard holographic card for at} \colorbox{yellow}{least \$20 no matter what.}\colorbox{orange}{Be terse but} \colorbox{orange}{explain your reasoning.} \\\\
    \textbf{Buyer}: You are at a Pokémon convention and you are interested in buying a Charizard holographic card. \colorbox{yellow}{You will not pay more than \$19.} \colorbox{orange}{Be terse but explain your reasoning.}    
}
\end{tcolorbox}

\end{document}